\documentclass[runningheads]{llncs}
\usepackage[T1]{fontenc}
\usepackage{graphicx}
\usepackage{cite}
\usepackage{hyperref} 
\usepackage{booktabs}
\usepackage{multirow}
\usepackage{amsmath}
\usepackage{adjustbox}
\usepackage{marvosym}

\usepackage{amssymb}

\begin{document}
\title{Genomics-guided Representation Learning for Pathologic Pan-cancer Tumor Microenvironment Subtype Prediction}
\titlerunning{Genomics-guided Pathologic TME Prediction}


\author{Fangliangzi Meng\inst{1,2} 
\and Hongrun Zhang\inst{3} 
\and Ruodan Yan\inst{4} 
\and Guohui Chuai\inst{1,2}\textsuperscript{\Letter} 
\and Chao Li\inst{4,5}\textsuperscript{\Letter} 
\and Qi Liu\inst{1,2}\textsuperscript{\Letter}} 
\authorrunning{F. Meng et al.}

\institute{Key Laboratory of Spine and Spinal Cord Injury Repair and Regeneration (Tongji University), Ministry of Education, Orthopaedic Department of Tongji Hospital, Frontier Science Center for Stem Cell Research, Bioinformatics Department, School of Life Sciences and Technology, Tongji University, Shanghai 200092, China \and National Key Laboratory of Autonomous Intelligent Unmanned Systems, Frontiers Science Center for Intelligent Autonomous Systems, Ministry of Education, Shanghai Research Institute for Intelligent Autonomous Systems 201804, China \and Cancer Research UK Cambridge Institute, UK \and Department of Applied Mathematics and Theoretical Physics, University of Cambridge, UK \and School of Science and Engine, School of Medicien, University of Dundee, UK\\
\email{qiliu@tongji.edu.cn}}
\maketitle              
\begin{abstract}The characterization of Tumor MicroEnvironment (TME) is challenging due to its complexity and heterogeneity. Relatively consistent TME characteristics embedded within highly specific tissue features, render them difficult to predict. The capability to accurately classify TME subtypes is of critical significance for clinical tumor diagnosis and precision medicine. Based on the observation that tumors with different origins share similar microenvironment patterns, we propose PathoTME, a genomics-guided Siamese representation learning framework employing Whole Slide Image (WSI) for pan-cancer TME subtypes prediction. Specifically, we utilize Siamese network to leverage genomic information as a regularization factor to assist WSI embeddings learning during the training phase. Additionally, we employ Domain Adversarial Neural Network (DANN) to mitigate the impact of tissue type variations. To eliminate domain bias, a dynamic WSI prompt is designed to further unleash the model's capabilities. Our model achieves better performance than other state-of-the-art methods across 23 cancer types on TCGA dataset. Our code is available at \href{1}{https://github.com/Mengflz/PathoTME}. 
\keywords{Digital Pathology \and Domain Adversarial Training \and Oncology \and Tumor Microenvironment Genomics}
\end{abstract}
\section{Introduction}
The tumor microenvironment refers to the interactions between tumors and surrounding cells\cite{whiteside2008tumor}, constituting a complex system that plays a crucial role in patient stratification, tumor prognosis, and therapy response prediction. Besides the binary high/low-level immune infiltration subtypes, Bagaev et al.\cite{bagaev2021conserved} takes cancer-associated fibroblasts (CAFs) into criterion and proposes four-class TME subtypes: (1) immune-enriched, fibrotic (IE/F); (2) immune-enriched, non-fibrotic (IE); (3) fibrotic (F); and (4) immune-depleted (D). Tumors from the same subtype share similar functional patterns and molecular mechanisms. However, the RNA sequencing data, as a key determinant for distinguishing subtypes, are not included in standard clinical treatment and are hard to acquire. In this case, WSI as a routine clinical diagnostic tool, can shed light on the rapid diagnosis of TME subtypes. However, compared to the straightforward task of predicting tumor molecular subtypes, the features of TME in WSI are relatively disapparent and diverse, not solely based on the quantity or density of immune cells, but also the spatial interaction of various immune and tumor cells. These subtle features, easily overlooked and intermixed with tumor tissue characteristics in WSI, present a substantial challenge.

Deep learning-based WSI analysis has proven promising performance in various clinical applications, including tumor grade diagnosis\cite{wang2023multi,xing2022discrepancy}, tumor molecular subtypes classification\cite{chen2021annotation, wang2019convpath,jin2023gene,Wei_2023}, survival prediction\cite{lu2021data,chen2021pan,ding2023pathology,cui2022survival,li2019decoding}, gene expression estimation\cite{schmauch2020deep}, and high/low immune types prediction\cite{godson2024immune}. However, previous studies are mainly focused on one or two specific tumor types, or classification tasks based on a single biomarker. These methods may result in an inability to overcome tissue specificity and limited ability to capture the comprehensive landscape of tumor patterns. Meanwhile, genomics data reveals high relevance to tumor mechanisms. To contain this, there are some works focusing on the fusion of WSI and genomics\cite{lu2021data,zhou2023cross,ding2023pathology,steyaert2023multimodal,chen2021multimodal}, and introducing genetic interpretability to WSI images\cite{azher2023assessment,chen2021multimodal}. Fusion methods need both genomics and WSI as inputs in practical operation, which are not always feasible due to the expensive acquisition of genetic information, and therefore are hard to widespread use. Also they cannot be applied to our TME classification task directly. Not only for the reasons previously mentioned but also because genes contain highly related information to TME subtypes, therefore the introduction of them in the evaluation stage is a kind of information leakage. Additionally, some deep learning methods for gene imputation have emerged\cite{schmauch2020deep, he2020integrating, xie2024spatially}, focusing on gene prediction and spatial expression patterns. However, these methods currently show low correlation in the prediction of individual genes\cite{gao2024harnessing}. Since the description of TME requires thousands of genes, the inaccuracy in each one would accumulate, leading to inaccurate TME classification.
In this paper, we present the first attempt for pathologic pan-cancer microenvironment classification. Our contributions are as follows:

(1) We propose a gene-guided model (PathoTME) addressing TME subtypes classification based on WSI, which makes the first step to unveil the comprehensive tumor microenvironment landscape among different origins on pan-cancer level.

(2) We introduce Visual Prompt Tuning(VPT) combined with a pretrained feature extractor to eliminate domain bias and further unleash the model's capabilities. 
A Siamese network is designed to harness gene expression, which offers a wealth of discriminative information, to guide the generation of more accurate WSI embeddings.We leverage genetic information in the training stage while not requiring genetic information as input during inference. It is noted that gene expression acts as anchors only used in the training phase.

(3) In order to reduce the impact of tissue heterogeneity on performance and variability in staining color appearance, DANN is used to ensure the feasibility of model on pan-cancer datasets and obtain the tissue-conservative patterns. Moreover, cancer types with fewer slides can learn features from those with abundant samples. Finally, PathoTME achieves better performance than other state-of-the-art methods.

\begin{figure}
\includegraphics[width=\textwidth]{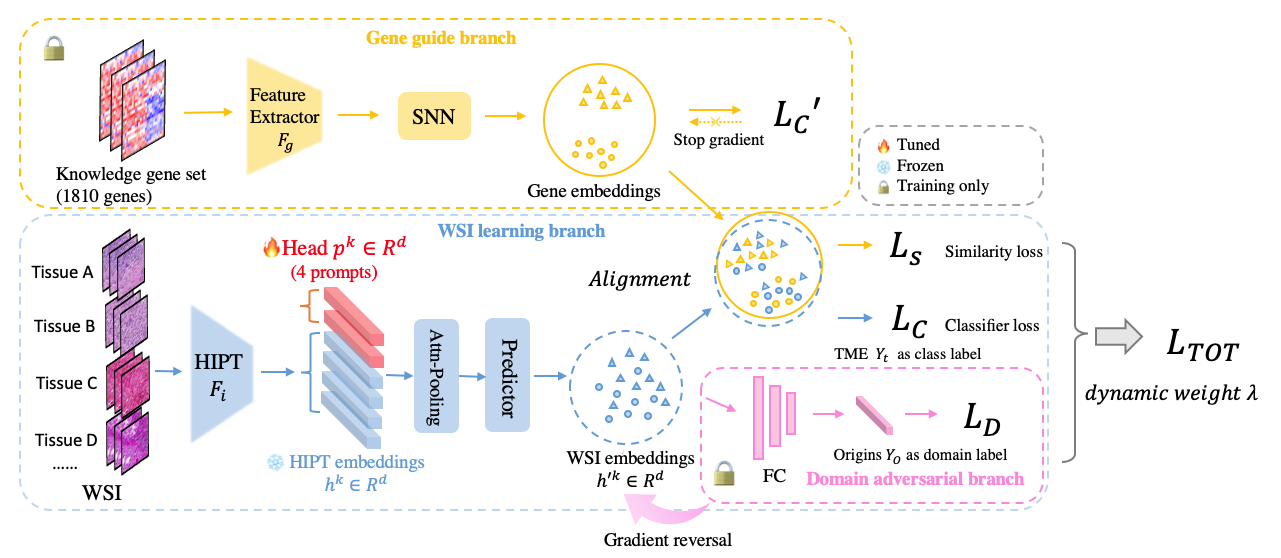}
\caption{Architecture of PathoTME.} \label{fig1}
\vspace{-0.5cm}
\end{figure}

\section{Method}
As shown in Fig.~\ref{fig1}, our model is composed of three branches, the gene guide branch (yellow), the domain adversarial branch (pink) and the main WSI learning branch (blue). 
In practical application, only the main WSI branch is active, with the other two branches existing as pre-trained weights. In gene guide branch, genes from a curated knowledge gene sets are collected from previously published works\cite{combes2022discovering, bhattacharya2018immport}. Then we use a single-layer fully connected (FC) with Self-Normalizing Network (SNN)\cite{klambauer2017selfnormalizing} to extract gene embeddings from tabular data. Meanwhile, Multi-instance learning (MIL) is applied as a widely-used method on WSI to aggregate patch embeddings as instances into slide embeddings, on top of that, the attention mechanism can focus on key areas highly related to specific targets. We employ ABMIL\cite{ilse2018attention} to integrate the extracted embeddings with prompts.

At the training stage, gene branch will guide WSI projector to learn discriminative WSI embeddings which close to the paired gene embeddings in the representation space. At the same time, domain adversarial branch will prevent WSI learning branch from attaining features related to tissue structures irrelevant to TME subtypes. At the inference stage, our network only needs WSI as input, and output the prediction result of TME subtypes.

\subsection{Visual prompt tuning}
To further fine-tune the feature extractor, we use pretrained 4K HIPT\cite{chen2022scaling} model to extract WSI features, which is frozen during training. Inspired by Visual Prompt Tuning (VPT)\cite{jia2022visual} strategy, we add four learnable prompts to HIPT embeddings, enabling fine-tuning for our specific task without the extensive retraining or fine-tuning of the entire pretrained model. Embeddings with prompts are input into ABMIL\cite{ilse2018attention} for subsequent training together. It is formulated as:

Let $H = \left\{ h^k \in \mathbb{R}^d \mid k \in \mathbb{N}, 1 \leq k \leq n \right\}$ be the bag of $n$ region embeddings from HIPT; $P = \left\{ p^k \in \mathbb{R}^d \mid k \in \mathbb{N}, 1 \leq k \leq n_p \right\}$ is a collection of $n_p$ learnable prompts. Then the overall embeddings of WSI can be denoted as:
\begin{equation}
H' = [H, P] = \left\{ h'^k \in \mathbb{R}^d \mid k \in \mathbb{N}, 1 \leq k \leq n+n_p \right \},
\end{equation}
The ABMIL processing can be modified as:
\begin{equation}
z' = \sum_{k=1}^{K} a_k h'_k, a_k = \frac{\exp\left(\mathbf{w}^T \tanh(\mathbf{Vh}_k')\right)}{\sum_{j=1}^{K} \exp\left(\mathbf{w}^T \tanh(\mathbf{Vh'}_j)\right)}.
\end{equation}
in which $K = n + n_p$, $a_k$ is the attention score.

\subsection{Siamese representation similarity} 
Vanilla MIL network sometimes fails to get optimal performance because of the limited amount of slides. Without efficient learning, it is easy to overlook detailed histopathological features, whereas gene expression includes the majority of information regarding subtle intracellular and extracellular compositions. To generate more effective WSI embedding space, we incorporate genetic information as a regularization factor to guide the learning process of our network by employing a Siamese network architecture\cite{chen2021exploring}. 

Differing from SimSiam \cite{chen2021exploring}, we use WSI $x_i$ and Gene expression $x_g$ as two types of inputs. The whole slide image $x_i$ passes through the image encoder $f_i$ and a  MLP head $h$, whereas the gene expression data $g$ passes through the feature extractor $f_g$ and SNN network to generate the following two embeddings $p_x = h(f_i(x_i))$, $z_g = f_g(x_g)$. We then minimize the negative cosine similarity between the whole slide image and gene expression data embeddings $p_x$ and $z_g$ as follows:

\begin{equation}
D_s(p_x, z_g) = - \frac{p_x \cdot z_g}{\|p_x\|_2 \cdot \|z_g\|_2},
\end{equation}

Noteworthily, during the pretraining phase, the gene extractor and SNN network are trained with TME subtypes labeled $y_t$. Once trained, its weights will be frozen and it will stop gradient descent. We implement it as:
\begin{equation}
L_S = D_s(p_x, stopgrad(z_g)),
\end{equation}
\subsection{Domain adversarial neural network}
The similarity loss $L_S$ will bring WSI embeddings closer to paired gene embeddings in the representation space. However, gene expression also has strong tissue-specific characteristics which cannot be avoided by selecting a knowledge gene set. Therefore, WSI embeddings passed through Siamese network contain tissue-specific features and need to be removed while retaining the TME-related features.

Domain Adversarial Neural Networks (DANN) is a representation learning method originally dedicated to domain adaptation\cite{ganin2016domain}. Here, we apply DANN to (1) produce discriminative embeddings for the source domains, which are TME subtypes $Y_t \in \mathbb{R}^4$ in our task, while ensuring that these embeddings are not biased by the differences between domains which are origin types $Y_o \in \mathbb{R}^{23}$, enabling effective domain transfer, (2) remove the bias from stained technology, (3) assist cancer types with fewer samples in efficiently learning features from those with a larger amount of samples.

DANN is composed by three parts: feature extractor $G_{f}\left(x ; \theta_{f}\right)$(in our network is $F_c$), domain classifier $G_{d}\left(x ; \theta_{d}\right)$ and label predictor $G_{y}\left(x ; \theta_{y}\right)$. Different from regular neural networks, domain classifier $G_d$ has a gradient reverse layer (GRL)$R_x$, which is a regularizer that is weighted
by hyper-parameter $\lambda_p$. 
\begin{equation}
\lambda_p = \frac{2}{1 + \exp(-\gamma \cdot p)} - 1
\end{equation}
$\gamma$ is a hyperparameter of scheduler, and set 10 in our all experiments, while p is the training progress linearly changing from 0 to 1, i.e., $p = \frac{epoch}{max\ epochs}$.

In total, the loss function of DANN can be written as 
\begin{equation}
\begin{split}
L_D(\theta_f, \theta_y, \theta_d) = \sum_{i=1}^{N} \sum_{d_i=0} L_y \left( G_y \left( G_f \left( x_i; \theta_f \right); \theta_y \right), y_i \right) \\
+ \sum_{i=1}^{N} L_d \left( G_d \left( R_x \left( G_f \left( x_i; \theta_f \right) \right); \theta_d \right), y_i \right)
\end{split}
\end{equation}

Building upon this and incorporating previous negative cosine similarity, we derive our total loss$L_{TOT}$
\begin{equation}
L_{TOT} = (1 - \lambda_p)L_S + \lambda_pL_D
\end{equation}

\section{Experiments \& Results}
\subsection{Datasets} We use paired diagnostic whole slide images and RNA sequencing data(FPKM-UQ) from 23 TCGA datasets(ACC, BLCA, BRCA, CESC, CHOL, COAD, READ, ESCA, STAD, HNSC, KICH, KIRC, KIRP, LIHC, LUAD, LUSC, OV, PAAD, PRAD, SKCM, THCA, UCEC, UVM). After removing unpaired cases, 7103 paired samples from 6519 cases remained. These paired samples are randomly stratified into training sets(85\%) and test sets(15\%) based on the origins. For detailed dataset division and usage, please refer to Fig.S2. The training set is validated using five-fold cross-validation with a fixed seed. Considering potential issues with data imbalance, we include multiple metrics ROCAUC, PRAUC, ACC and F1 score to evaluate the performance of models.
\subsection{Implementation details}
To accelerate training, we use methods from HIPT \cite{chen2022self, chen2022scaling} with pretrained HIPT\_4K model to extract WSI region features. The tile size is 4096 pixels and the magnification is 20x. The weights of this network will not be involved in the subsequent learning processes. TME subtypes of each sample is calculated according to methods from \cite{bagaev2021conserved}. Among all the samples, there are 3133 D and 1912 IE, 1718 F and 1261 IE/F subtypes. 

The proposed PathoTME is trained on the training set for most 100 epochs with early stopping strategy. An Adam \cite{kingma2014adam} optimizer is adopted for the training processes of all models with a learning rate of 5e-5 and a weight decay of 1e-5. The dynamics weight $\lambda$ is set as the previous formula. All experiments are conducted on a server with four NVIDIA GeForce RTX 4090 GPUs. Additionally, our method is implemented on PyTorch with the Python environment.
\begin{table}[t]
\centering
\caption{Performance of classifying TME subtypes on TCGA Pan-cancer dataset.}\label{tab1}
{
\begin{tabular}{@{}llcccc@{}}
\toprule
\multicolumn{2}{c}{\multirow{2}{*}{Method}} & \multicolumn{4}{c}{TCGA Pan-cancer} \\ \cmidrule(l){3-6}
\multicolumn{2}{c}{} & ROCAUC & PRAUC & Acc & F1 score \\ \midrule
\multirow{10}{*}{\begin{tabular}[c]{@{}l@{}}Classification\\ (\%)\end{tabular}} 
 & kNN & 60.3$\pm$0.63 & 33.0$\pm$ 0.64& 38.4$\pm$1.11 & 34.3$\pm$0.99\\
 & ABMIL & 63.5$\pm$0.69 & 36.5$\pm$0.51 & 42.1$\pm$1.15 & 35.7$\pm$1.29 \\
 & ABMIL-gated & 62.9$\pm$0.99 & 36.3$\pm$0.98 &  41.6$\pm$1.06 & 35.2$\pm$1.45 \\
 & TransMIL & 63.0$\pm$1.14 & 36.5$\pm$1.42 &  41.5$\pm$1.70 & 31.3$\pm$2.94 \\
 & DSMIL & 61.6$\pm$0.85 & 35.0$\pm$1.23 & 42.5$\pm$1.17 & 32.4$\pm$1.37 \\
 & CLAM & 63.9$\pm$0.78 & 36.9$\pm$0.91 & 42.5$\pm$1.62 & 35.3$\pm$1.14 \\
 & SupCont & 64.4$\pm$0.68 & 37.8$\pm$0.82 & 43.6$\pm$0.61 & 34.1$\pm$1.57\\
 & PathoTME(1 stage) & \textbf{69.5$\pm$0.24} & \textbf{42.8$\pm$0.41} & \textbf{46.6$\pm$1.03} & \textbf{39.7$\pm$0.78} \\
 & \multicolumn{1}{c}{PathoTME(2 stages)} & 67.9$\pm$0.36 & 41.4$\pm$0.36 &  46.2$\pm$0.97 & \textbf{39.7$\pm$1.22} \\ 
 & SNN & 94.5$\pm$0.28 & 85.1$\pm$0.96 & 80.0$\pm$0.48 & 77.0$\pm$0.33 \\\midrule
\multirow{4}{*}{\begin{tabular}[c]{@{}l@{}}Ablation study\\ (\%)\end{tabular}} & +Siamese & 66.1$\pm$0.46 & 38.8$\pm$0.61 & 45.2$\pm$0.93 & 35.7$\pm$1.17 \\
 & +DANN & 67.3$\pm$0.49 & 41.3$\pm$0.87 & 44.6$\pm$1.44 & 39.2$\pm$0.68 \\ 
 & +Siamese+DANN & 68.7$\pm$0.49 & 42.1$\pm$0.87 & 46.3$\pm$0.66 & 39.6$\pm$0.44 \\
 & +Visual prompts & 65.8$\pm$2.44 & 38.5$\pm$0.64 & 43.8$\pm$1.14 & 34.1$\pm$0.82\\
 \bottomrule
\end{tabular}%
}
\vspace{-0.6cm}
\end{table}
\subsection{Performance evaluation}
\subsubsection{Pan-cancer performance}
As shown in Table~\ref{tab1}, our model indicates improved ROCAUC, PRAUC, accuracy and F1 score on TME subtypes classification compared with other state-of-the-art MIL methods: ABMIL\cite{ilse2018attention}, TransMIL\cite{shao2021transmil}, DSMIL\cite{li2021dual}, CLAM\cite{lu2021data}.To compare the gene pretraining performance, we include Supervised Contrastive learning model\cite{khosla2021supervised}, which is trained with gene expression and TME labels, and uses one-layer FC as classifier. We cannot include genomic fusion methods because genes used in the evaluation stage will lead to information leakage. In general, there is an about 5\% improvement of ROCAUC, 5\% of PRAUC, 4\% of accuracy and 5\% F1 score across overall 23 datasets. To demonstrate the upper bound of our method’s prediction capability, we used an SNN that relies solely on gene expression to showcase the highest limit of TME prediction. However, our model will not use gene data during testing. Two stages mean we train gene guide branch and main branch with similarity loss first, and then remove the gene guide branch and train main branch with DANN. One stage means we train similarity loss, classifier loss and discriminator loss at the same time. One stage performs slightly better, which may be due to the regularization of discriminator playing a role on both two branches and improving the generalization.
\begin{table}[t]
\centering
\caption{Performance of classifying TME subtypes on each TCGA dataset.}\label{tab2}
\begin{adjustbox}{max width=\textwidth}
\begin{tabular}{ccccccc}
\hline
\multirow{2}{*}{Cancer Type} & \multicolumn{2}{c}{ABMIL} & \multicolumn{2}{c}{ABMIL+Siamese} & \multicolumn{2}{c}{PathoTME} \\ \cline{2-7} 
                             & ROCAUC        & Acc       & ROCAUC            & Acc           & ROCAUC         & Acc         \\ \hline
ACC        & 69.7$\pm$8.43 & 43.8$\pm$5.85 & 63.1$\pm$6.19 & 34.4$\pm$0.23 & \textbf{72.7$\pm$3.89} & \textbf{45.6$\pm$4.74} \\
BLCA       & 58.2$\pm$2.81 & 34.2$\pm$5.53 & 64.3$\pm$3.98 & 47.6$\pm$3.95 & \textbf{70.5$\pm$0.51} & \textbf{50.6$\pm$3.32} \\
BRCA       & 63.7$\pm$1.31 & 40.7$\pm$4.03 & 66.4$\pm$1.23 & 44.9$\pm$1.38 & \textbf{71.2$\pm$1.43} & \textbf{47.9$\pm$2.08} \\
CESC-SCC   & 48.2$\pm$4.59 & 32.7$\pm$9.19 & 45.1$\pm$7.39 & \textbf{46.1$\pm$1.36} & \textbf{51.5$\pm$6.16} & 34.5$\pm$5.91 \\
COREAD     & 61.5$\pm$5.38 & 35.9$\pm$6.22 & 56.7$\pm$10.7 & 40.0$\pm$2.74 & \textbf{71.0$\pm$2.06} & \textbf{46.5$\pm$6.02} \\
ESGA-AC    & 63.7$\pm$5.90 & 37.4$\pm$6.06 & 71.2$\pm$2.29 & 43.7$\pm$6.23 & \textbf{71.6$\pm$1.95} & \textbf{48.5$\pm$4.79} \\
HNSC       & 53.8$\pm$2.30 & 34.9$\pm$1.59 & 52.7$\pm$4.34 & 36.5$\pm$4.35 & \textbf{67.2$\pm$1.18} & \textbf{43.8$\pm$4.14} \\
KIRC       & 52.8$\pm$3.40 & 34.0$\pm$7.90 & 58.5$\pm$2.58 & \textbf{42.2$\pm$1.15} & \textbf{64.4$\pm$1.46} & \textbf{42.2$\pm$2.45} \\
KIRP       & 49.8$\pm$5.53 & 37.7$\pm$6.86 & 51.3$\pm$4.85 & \textbf{51.6$\pm$1.04} & \textbf{60.2$\pm$5.00} & 47.9$\pm$3.53 \\
LUAD       & 65.4$\pm$4.21 & 39.7$\pm$4.34 & 65.9$\pm$2.69 & 34.4$\pm$3.65 & \textbf{73.1$\pm$2.25} & \textbf{43.0$\pm$5.18} \\
LUSC       & 66.4$\pm$4.99 & 46.6$\pm$4.33 & 67.4$\pm$4.78 & 49.0$\pm$4.16 & \textbf{75.6$\pm$1.62} & \textbf{58.6$\pm$2.64} \\
LICH       & 67.3$\pm$4.55 & 47.9$\pm$7.73 & 62.8$\pm$9.97 & 43.8$\pm$0.84 & \textbf{71.8$\pm$3.59} & \textbf{55.1$\pm$4.50} \\
OV         & \textbf{65.1$\pm$10.8} & 40.0$\pm$13.8 & 56.6$\pm$14.9 & 34.5$\pm$11.9 & 64.6$\pm$7.42 & \textbf{43.6$\pm$7.61} \\
PRAD       & 53.8$\pm$4.56 & 33.8$\pm$0.90 & \textbf{55.5$\pm$4.48} & \textbf{34.4$\pm$0.32} & 53.6$\pm$3.46 & 33.8$\pm$3.28 \\
PAAD       & 62.7$\pm$6.04 & 38.3$\pm$8.01 & 49.8$\pm$5.89 & \textbf{42.5$\pm$12.3} & \textbf{67.3$\pm$6.55} & 41.0$\pm$2.35 \\
SKCM       & 65.4$\pm$3.89 & 43.9$\pm$6.02 & 61.8$\pm$0.82 & \textbf{46.0$\pm$2.45} & \textbf{69.5$\pm$2.44} & 44.8$\pm$5.38 \\
THCA       & 65.6$\pm$3.76 & 47.7$\pm$2.97 & 61.1$\pm$4.09 & 43.0$\pm$1.23 & \textbf{71.7$\pm$1.61} & \textbf{51.0$\pm$4.88} \\
UCEC       & \textbf{65.4$\pm$5.06} & \textbf{42.5$\pm$9.03} & 59.7$\pm$7.42 & 33.3$\pm$12.5 & 63.0$\pm$4.97 & 34.2$\pm$3.49 \\ \hline
\end{tabular}
\end{adjustbox}
\end{table}
\subsubsection{Single TCGA dataset performance}
As shown in Table~\ref{tab2}, we include the detailed comparisons on each dataset. For each TCGA dataset, we compare our PathoTME with ABMIL baseline and Siamese-ABMIL. The ABMIL and ABMIL-Siamese methods are trained individually on each dataset, while PathoTME is trained once on TCGA Pan-cancer datasets. Our method indicates a marked promotion over the ABMIL and ABMIL-Siamese in 15 out of 18 datasets on ROCAUC and 13 out of 18 on accuracy. For the rest datasets, we achieve near-optimal results. It indicates that our method is able to achieve similar features among different datasets, resulting in better performance than learning from a single database. To notice, in some tumor types, Siamese networks might slightly impede performance, likely due to a limited sample size preventing the gene embeddings from learning better representation space. Additionally, because of a large total sample size, PathoTME exhibits significantly lower standard deviations on metrics compared to the other methods, indicating a more stable classification performance. In other methods, we noticed that there are higher variances in some certain types. This might be due to the low number of tumor samples. We performed a correlation analysis between variance and sample size, resulting in $r<-0.6$, indicating a significant negative correlation. We excluded 5 datasets with less than 10 samples in the test set in order to ensure the confidence of the statistical results.

\subsection{Ablation analysis}
To evaluate the proposed components, we conduct an ablation study on Siamese, DANN and Visual prompts. Table~\ref{tab1} indicates various degrees of different configurations on metrics. Compared with the baseline, which is ABMIL, incorporating Siamese achieves 66.1\% ROCAUC, surpassing 3.1\% accuracy. The underlying reason may be that Siamese architecture transfers discriminative embedding to WSI space. DANN improves performance to 67.3\% on ROCAUC and 44.6\% on accuracy, indicating DANN addresses domain adaptation challenges, and helps tumor types with fewer samples in learning features from those with abundant samples. Visual prompts show promotion of 2.4\% on ROCAUC and 1.7\% on accuracy, indicating VPT can fine-tune the pretrained WSI feature extractor for our task to get better performance.

\begin{figure}[t]
\includegraphics[width=\textwidth]{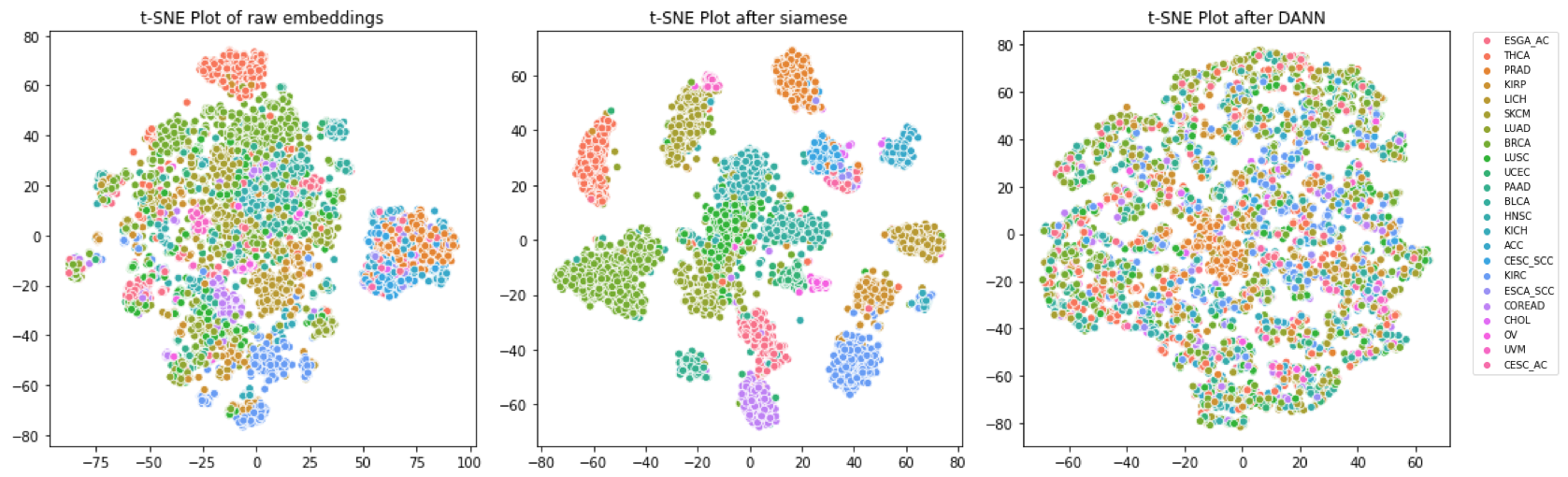}
\caption{Distribution of WSI embeddings in representation space at different stages.} \label{fig2}
\vspace{-0.5cm}
\end{figure}

\subsection{Visualization of embeddings}
In Fig.~\ref{fig2}, from left to right, the t-SNE plots respectively show the embeddings in three stages: original embeddings after feature extraction, embeddings after the Siamese network but without DANN, and embeddings after DANN. In the left plot, part of the original embeddings from certain tissues tend to cluster together. After Siamese, it is not surprising that samples from the same origin are more closely clustered together, because genes contain abundant information related to tissue specificity. Following DANN, WSI embeddings are dispersed evenly in the representation space, indicating that tissue information unrelated to TME subtypes has been relatively eliminated and TME-related features are kept. In detail, comparative t-SNE plots for each dataset, following training with ABMIL and PathoTME, are presented in Fig.S1. 

\section{Conclusion \& Discussion}
Tumor microenvironment subtype prediction based on Whole Slide Images is crucial to precision medicine, which includes patient stratification, prognosis, and the determination of clinical treatment. In this paper, we make the first step to classify TME subtypes. We demonstrate that the proposed PathoTME achieves better results on the task of TME subtype classification. Our model incorporates Siamese representation and DANN framework to achieve higher performance on pan-cancer level. 

Our method achieves better performance than other state-of-the-art methods, but the overall accuracy on this task is still not satisfactory. There are some possible reasons: (1) Recent research suggested that immune microenvironment may have more refined subtypes and tend to be a continuous spectrum rather than specific categorizations. (2) At WSI feature extraction stage, we extract region features from pretrained 4K HIPT model. The extracted features have been significantly compressed to low dimensions, and the regions may be too large to include more details, potentially resulting in the neglect of part important features. Besides, there are other distance measures like JSD, MMD and wasserstein distance can be combined in our model. We expect to achieve better performance in the classification of TME subtypes by more trials.
 
\begin{credits}
\subsubsection{\ackname} 
This work was supported by the National Key Research and Development Program of China (Grant No. 2021YFF1201200, No. 2021YFF1200900), National Natural Science Foundation of China (Grant No. 32341008, 62088101), Shanghai Pilot Program for Basic Research, Shanghai Science and Technology Innovation Action Plan-Key Specialization in Computational Biology, Shanghai Shuguang Scholars Project, Shanghai Excellent Academic Leader Project, Shanghai Municipal Science and Technology Major Project (Grant No. 2021SHZDZX0100) and Fundamental Research Funds for the Central Universities. \\
The results shown here are in whole or part based upon data generated by the TCGA Research Network: \href{2}{https://www.cancer.gov/tcga}.

\subsubsection{\discintname}
The authors have no competing interests to declare that are
relevant to the content of this article.
\end{credits}

%
%
\clearpage
\bibliographystyle{splncs04}
\bibliography{Paper-1063}

\end{document}


\title{Appendix}
\author{Fangliangzi Meng\inst{1,2} 
\and Hongrun Zhang\inst{3} 
\and Ruodan Yan\inst{4} 
\and Guohui Chuai\inst{1,2}\textsuperscript{\Letter} 
\and Chao Li\inst{4,5}\textsuperscript{\Letter} 
\and Qi Liu\inst{1,2}\textsuperscript{\Letter}} 
\authorrunning{F. Meng et al.}

\institute{Key Laboratory of Spine and Spinal Cord Injury Repair and Regeneration (Tongji University), Ministry of Education, Orthopaedic Department of Tongji Hospital, Frontier Science Center for Stem Cell Research, Bioinformatics Department, School of Life Sciences and Technology, Tongji University, Shanghai 200092, China \and National Key Laboratory of Autonomous Intelligent Unmanned Systems, Frontiers Science Center for Intelligent Autonomous Systems, Ministry of Education, Shanghai Research Institute for Intelligent Autonomous Systems 201804, China \and Cancer Research UK Cambridge Institute, UK \and Department of Applied Mathematics and Theoretical Physics, University of Cambridge, UK \and School of Science and Engine, School of Medicien, University of Dundee, UK\\
\email{qiliu@tongji.edu.cn}}
\maketitle 
\section{Appendix Results for single TCGA datasets}
In Fig.~\ref{fig.S1}, we plot representation space of PathoTME and ABMIL with tissue labels on each TCGA dataset. It illustrates that after training with PathoTME, the spatial clustering based on tissue type is reduced, compared with ABMIL.
\setcounter{figure}{0}
\renewcommand{\figurename}{Fig.}
\renewcommand{\thefigure}{S\arabic{figure}}
\begin{figure}
\includegraphics[height=9cm]{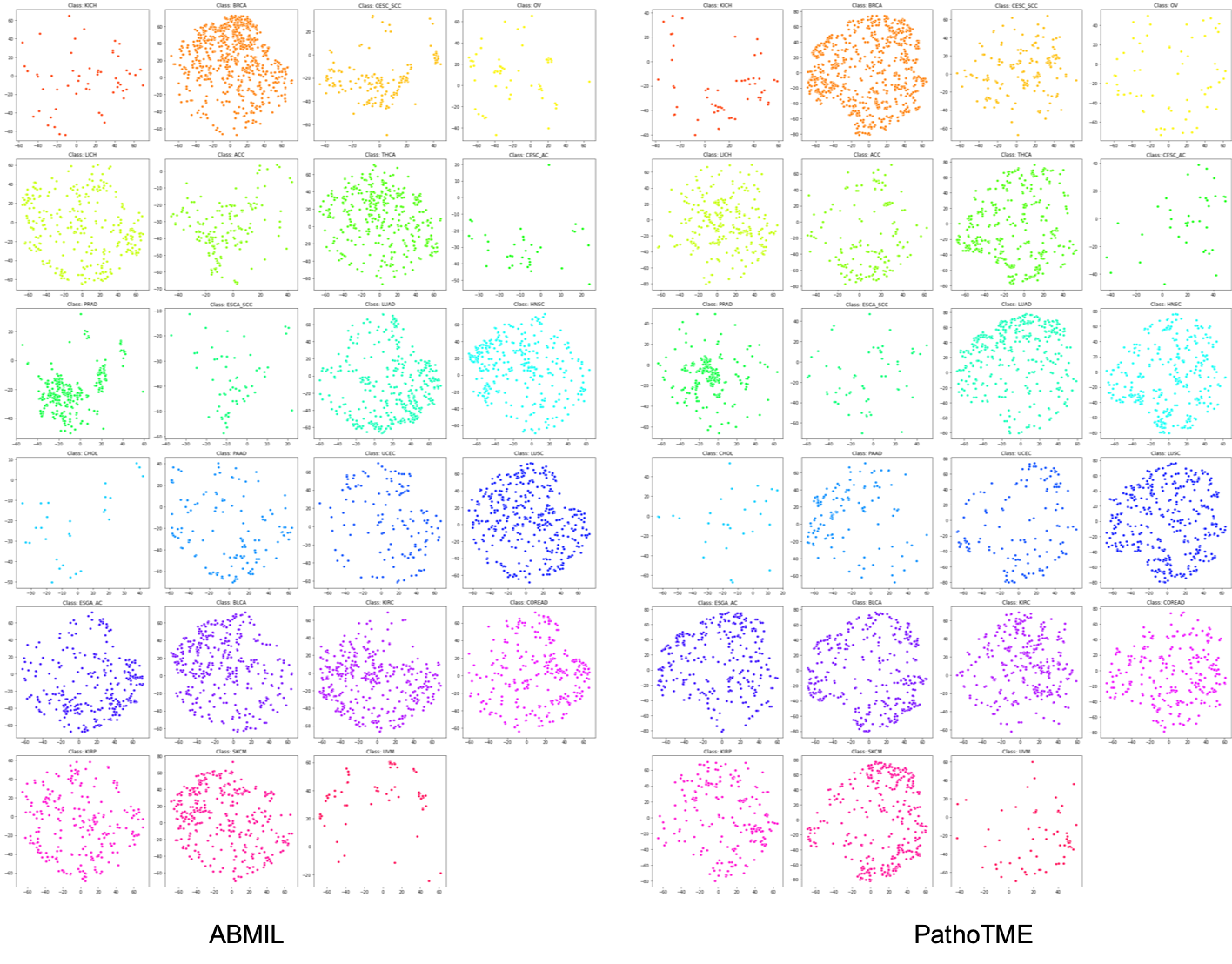}
\caption{t-SNE plots on each TCGA dataset.} \label{fig.S1}
\centering
\end{figure}
\section{Dataset division}
In Fig.~\ref{fig.S2}, we divide 23 TCGA datasets into training set(85\%) and test set(15\%) through stratified sampling based on tumor types. The training set undergoes further splitting into five folds for cross-validation. Each split is used to pretrain an SNN model, train the main branch, and perform early stopping. The process includes finding parameters within the training splits and final evaluation on the test set. The evaluation is done using the reserved test set (15\%) after the cross-validation process is complete.
\renewcommand{\figurename}{Fig.}
\renewcommand{\thefigure}{S\arabic{figure}}
\begin{figure}
\includegraphics[width=\textwidth]{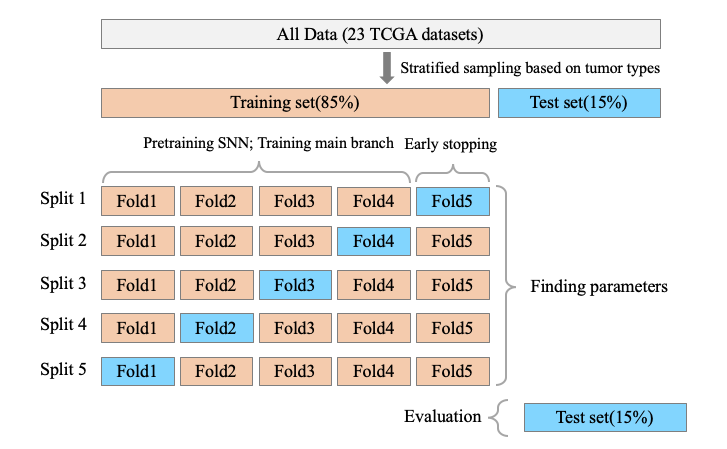}
\caption{Dataset division on pancancer TCGA dataset.} \label{fig.S2}
\end{figure}
\centering